\documentclass[3p]{elsarticle} 
\usepackage[hyphens]{url}

\makeatletter 
\def\ps@pprintTitle{ 
\let\@oddhead\@empty 
\let\@evenhead\@empty 
\def\@oddfoot{\hfill\thepage} 
\def\@evenfoot{\thepage\hfill}}
\makeatother

\usepackage{lineno} 
\providecommand{\tightlist}{%
  \setlength{\itemsep}{0pt}\setlength{\parskip}{0pt}}

\biboptions{sort&compress} 
\usepackage{graphicx}
\usepackage{booktabs} 

\usepackage[T1]{fontenc}
\usepackage{lmodern}
\usepackage{amssymb,amsmath}
\usepackage{ifxetex,ifluatex}
\usepackage{fixltx2e} 
\IfFileExists{upquote.sty}{\usepackage{upquote}}{}
\ifnum 0\ifxetex 1\fi\ifluatex 1\fi=0 
  \usepackage[utf8]{inputenc}
\else 
  \usepackage{fontspec}
  \ifxetex
    \usepackage{xltxtra,xunicode}
  \fi
  \defaultfontfeatures{Mapping=tex-text,Scale=MatchLowercase}
  
\fi
\IfFileExists{microtype.sty}{\usepackage{microtype}}{}
\usepackage{longtable}
\usepackage{graphicx}
\makeatletter
\def\maxwidth{\ifdim\Gin@nat@width>\linewidth\linewidth
\else\Gin@nat@width\fi}
\makeatother
\let\Oldincludegraphics\includegraphics
\renewcommand{\includegraphics}[1]{\Oldincludegraphics[width=\maxwidth]{#1}}
\ifxetex
  \usepackage[setpagesize=false, 
              unicode=false, 
              xetex]{hyperref}
\else
  \usepackage[unicode=true]{hyperref}
\fi
\hypersetup{breaklinks=true,
            bookmarks=true,
            pdfauthor={},
            pdftitle={Unique Metric for Health Analysis with Optimization of Clustering Activity and Cross Comparison of Results from Different Approach},
            colorlinks=true,
            urlcolor=blue,
            linkcolor=magenta,
            pdfborder={0 0 0}}
\usepackage{lipsum}

\begin{document}
\begin{frontmatter}

  \title{Unique Metric for Health Analysis with Optimization of Clustering
Activity and Cross Comparison of Results from Different Approach}
    \author[Researchers]{Kumarjit Pathak}
   \ead{Kumarjit.pathak@outlook.com} 
  
    \author[Researchers]{Jitin Kapila}
   \ead{Jitin.kapila@outlook.com} 
  
      \address[Researchers]{Data Science Researchers, Bangalore, India}
  
  \begin{abstract}
  In machine learning and data mining, Cluster analysis is one of the most
  widely used unsupervised learning technique. Philosophy of this
  algorithm is to find similar data items and group them together based on
  any distance function in multidimensional space. These methods are
  suitable for finding groups of data that behave in a coherent fashion.
  The perspective may vary for clustering i.e.~the way we want to find
  similarity, some methods are based on distance such as K-Means technique
  and some are probability based, like GMM. Understanding prominent
  segment of data is always challenging as multidimension space does not
  allow us to have a look and feel of the distance or any visual context
  on the health of the clustering.
  
  While explaining data using clusters, the major problem is to tell how
  many cluster are good enough to explain the data. Generally basic
  descriptive statistics are used to estimate cluster behaviour like scree
  plot, dendrogram etc. We propose a novel method to understand the
  cluster behaviour which can be used not only to find right number of
  clusters but can also be used to access the difference of health between
  different clustering methods on same data. Our technique would also help
  to also eliminate the noisy variables and optimize the clustering
  result.
  \end{abstract}
  
 \end{frontmatter}

\hypertarget{introduction}{%
\section{1. Introduction}\label{introduction}}

\quad

Unsupervised learning is part of machine learning, where the objective
of the methods is to understand the patterns or classes without any
supervision or pre-defined labels. Clustering techniques are few of the
important methods to achieve these objectives. In last few decades there
has been great momentum in using and advancing these methods to make
sense from the data.

These methods proved to be a great tool for understanding and creating
groups of coherent behaviours within the data. Immense business
importance of these unsupervised techniques has helped this to become
continuous research topic. It helps in identifying different customer
cohorts based on different attributes and help companies to make the
right targeting policy customized for each group of customers.

Seeking such sense from data in right fashion is one of the daunting
task for statisticians and business users. Current practice to
understand the similarity and dis-similarity between the clusters are
heavily dependent on descriptive statistics like average, median and IQR
etc.

It becomes even more challenging when statisticians try to compare
different type of clustering methods and see which method is better than
other.

In this paper we propose a novel approach which is not any clustering
method specific and can be applied to all clustering methods. Our metric
is only dependent on the number of observations.

The rest of paper is organised as follows. In section 2 we deep dive on
literature survey. In Section 3 we propose our formulation. Section 4
explains the procedure in detail. Experiment and results are discussed
in Section 5. Concluding remarks is mentioned in section 6.

\hypertarget{literature-review}{%
\section{2. Literature Review}\label{literature-review}}

\quad

Every clustering method has its own unique perspective towards the data.
K-means and other distance-based methods look from distance similarity
and closeness approximates in multidimensional space. Whereas methods
like Gaussian Mixture Models (GMM) looks at populistic space similarity.
This is summarized in Table- 1 which contains most used methods and
underlying similarity reference they use to understand the data
segments.

\quad

\begin{longtable}[]{@{}lc@{}}
\caption{Clusering methods and measures}\tabularnewline
\toprule
\begin{minipage}[b]{0.37\columnwidth}\raggedright
Underlying Measure\strut
\end{minipage} & \begin{minipage}[b]{0.57\columnwidth}\centering
Clustering Method\strut
\end{minipage}\tabularnewline
\midrule
\endfirsthead
\toprule
\begin{minipage}[b]{0.37\columnwidth}\raggedright
Underlying Measure\strut
\end{minipage} & \begin{minipage}[b]{0.57\columnwidth}\centering
Clustering Method\strut
\end{minipage}\tabularnewline
\midrule
\endhead
\begin{minipage}[t]{0.37\columnwidth}\raggedright
Distances / Similarities\strut
\end{minipage} & \begin{minipage}[t]{0.57\columnwidth}\centering
K-Means and variants, Hierarchical Clustering, Kernel PCA, Local Linear
Embeddings, ISOMAPS, t-SNE, etc.\strut
\end{minipage}\tabularnewline
\begin{minipage}[t]{0.37\columnwidth}\raggedright
Probability\strut
\end{minipage} & \begin{minipage}[t]{0.57\columnwidth}\centering
Latent Dirichlet Allocation, Gaussian Mixture Models, Probabilistic
Trees, etc\strut
\end{minipage}\tabularnewline
\begin{minipage}[t]{0.37\columnwidth}\raggedright
Information / Criterion\strut
\end{minipage} & \begin{minipage}[t]{0.57\columnwidth}\centering
Birch, Self-Organizing Feature Map, Density linkage-based methods
(DBSCAN and variants), NMF-EM, etc.\strut
\end{minipage}\tabularnewline
\bottomrule
\end{longtable}

A detailed view on these ar given by Xu and Wunsch (2005) and Berkhin
(2002).

\quad

There has been great deal of work done in assessing different measures
by O. Arbelaitz (2013), H. Chouikhi (2015), H. Meroufel (2017) and J.
Hamalainen (2017). Most of these measures are readily available for
experimentation. Though being statistically sound they are all based on
certain assumptions about the clustering method being used. We postulate
a methodology invariant technique to compare cluster health. Broadly
existing cluster evaluation can be defined on following criterions:

\begin{itemize}
\tightlist
\item
  Different Clustering algorithms
\item
  Data or types of Data on which Clustering is being applied
\item
  Boundary definitions given by Clustering method
\item
  Pre-supposed class similarity or Class based behaviour within Cluster
\item
  Repeatability of Experiments
\item
  Explainability of Clusters
\end{itemize}

\hypertarget{different-clustering-algorithmss}{%
\paragraph{Different Clustering
algorithmss:}\label{different-clustering-algorithmss}}

\quad

Different clustering methods are defined by its similarity measurement
methodologies. Different methods use different underlying measures to
access the data coherence and hence evaluation and comparison of the
intra-cluster health does not have a common base. Often different
methods of clustering contradict with each other. This is a challenge to
compare a distance-based clustering output vs a probability-based
clustering. For example, A cluster with very small within-cluster
distance in k-Means can have absolute dissimilarity of grouping in GMM
and may not fall in the same bucket. In this case if statistician needs
to take a decision which clustering is better for grouping is a real
challenge. There is no direct method to access this yet. It becomes very
specific to data, domain knowledge and objective of clustering

\hypertarget{data-or-types-of-data-on-which-clustering-is-being-applied}{%
\paragraph{Data or types of Data on which Clustering is being
applied:}\label{data-or-types-of-data-on-which-clustering-is-being-applied}}

\quad

Data can be in varied formats. While most methods accept mixed data
formats (Categorical and numeric), it is always a decision of
statistician to make data have business sense.

\hypertarget{boundary-definitions-given-by-clustering-method}{%
\paragraph{Boundary definitions given by Clustering
method:}\label{boundary-definitions-given-by-clustering-method}}

\quad

The main task of clustering is to group or segregate the data in a way
where it explains certain common behaviour within clusters. To access
that, following measures are currently available based on our literature
review , including work from M. K. Pakhira (2004):

\begin{enumerate}
\def\labelenumi{\arabic{enumi})}
\tightlist
\item
  Silhoutte Index
\item
  Davies - Bouldin Index
\item
  Dunn Index
\item
  Partition Coefficient
\item
  Separation Index
\item
  Xie - Benie Index
\item
  Ratkowsky - Lance ratio
\item
  Goodman - Kruskal Gamma
\item
  Hubert - Levin (C - Index)
\item
  Krzanowski - Lai Index
\item
  Pakhira - Bandyopadhyay - Maulik(PBM)
\item
  Wemmert - Gancarsk Index
\item
  Ray - Turi Index
\end{enumerate}

\quad

Though these extensive measures are there but being very subjective to
clustering method. This act as the short comings and hence cannot be
used to access cross cluster comparison.

\hypertarget{pre-supposed-class-similarity-or-class-based-behaviour-within-cluster}{%
\paragraph{Pre-supposed class similarity or Class based behaviour within
Cluster:}\label{pre-supposed-class-similarity-or-class-based-behaviour-within-cluster}}

\quad

Like boundary definitions, class-based comprehension has been another
significant way evaluation metrics are being used to understand
clusters. Though assessing cluster, based on predefined labels or class
is not ideal, as it defeats the whole purpose of unsupervised learning.
But still these labels are good to explain the behaviour of cluster
which remain outside or unconsumed in clustering activity. One good way
to use this is leave a categorical variable out of clustering methods
and compare all approaches of clustering based on this left out variable
or variable set. Few of these metrics are:

\begin{enumerate}
\def\labelenumi{\arabic{enumi})}
\tightlist
\item
  Accuracy
\item
  F-measure
\item
  Normalized Mutual Information
\item
  Rand Index
\item
  Alternative Dunn Index
\item
  Fowlkes - Mallows Index
\item
  Dice Index
\item
  V - Measure
\item
  Entropy and Purity
\end{enumerate}

\hypertarget{repeatability-of-experiments}{%
\paragraph{Repeatability of
Experiments:}\label{repeatability-of-experiments}}

\quad

Another way to access clusters is to see whether same group of data
occurs consistently within a cluster repeatedly. This behaviour is only
possible when the data is coherent and shows relations with the label
and not grouped by mere randomness in data. This is often evaluated by
cross validation and re-iterations. The major problem of such evaluation
is that one might need to calibrate the cluster group as ``cluster
name'', which may vary over iterations and hence changes during
iterations may not be comparable. This becomes tedious when number of
cluster increases and data in not actually less coherent.

\hypertarget{explainability-of-clusters}{%
\paragraph{Explainability of
Clusters:}\label{explainability-of-clusters}}

\quad

The main purpose of this type of evaluation is to make ``sense'' of
clusters. This evaluation is more business driven than statistical.
Explainable clusters are difficult to find but if found they might have
a huge impact on analysis of data. There is no standard method for such
evaluation and it majority of times depends on domain knowledge and
manual comprehension.

\quad

From all the above study few things that can be noted for defining a
``good'' \(metric\) or a concrete clustering criterio for cluster
evalutions should posses following properties:

\begin{enumerate}
\def\labelenumi{\arabic{enumi})}
\tightlist
\item
  Should be invariant of method of clustering
\item
  Should be invariant of data type used to do clustering
\item
  Have capability to explain each variable used to do clustering
\item
  Should be evaluatable from external data / labels / classes
\item
  Should not be affected by different \emph{``name''} of cluster and
  represents true internal behaviour of cluster on data
\item
  Should be independent of any underlying measures used in clustering
\item
  Have consistence to repeated experiments
\end{enumerate}

Considering above points, we propose a method that systematically
comprehends clusters. These points provide us with the basic principles
/ conditions that metric should fulfil to cater the needs of unbiased
cluster evaluations.

\hypertarget{method}{%
\section{3. Method}\label{method}}

Our method creates a distribution of different quantile of each variable
vs each cluster. This multidimensional matrix is consumed for deriving
our metric for cluster health.

To understanding cluster behaviour, one might need to consider on how
much does the cluster groups each variable, and at same time it is very
important to understand how much data can be explained by the clusters.
Considering this we propose score \(S_v\) for \(k\) cluster w.r.t to
variable \(v\) as :

\begin{equation}
S_v^k = \frac{N_v^k}{max(l,k)} \ast ln(\frac{N_d}{l*k})
\label{eq:weigthed1}
\end{equation}

\quad

where first part of eqn. (1) is called the \(seggregation\ factor\) and
second part is called \(explaination\ factor\). The theory behind this
formulation is, that every cluster should be able to bucket / group /
comprehend each variable in a specific range of it's value which
idealistically should be different in any other cluster for the same
variable.

When we draw a cross-tab between unique interval of values of a variable
and k number of clusters we expect, Nd to be fill the matrices in such a
fashion that cross tab matrix \(M_v^k\) has(\(k\) = number of cluster ,
\(v\) = the variable under consideration) only diagonal values. The
maximum number of diagonal values possible is \(max(l,k)\) {[}\(l\) =
number of interval range considered for a variable, \(k\)= number of
clusters being evaluated{]}. Aiming to relax this segregation assumption
(diagonal distribution of range vs cluster), we consider the values
which are greater than \(median(M_v^k)\) of the frequency from the
matrix generated as segregated values and take sum of such instances in
\(M_v^k\) .

The ratio of segregated instances by least possible instances gives us
segregation factor. The reason for using \(median(M_v^k)\) and not any
other measure is that, other measures are usually influenced by the
range and mere occurrence of values in \(M_v^k\). Hence, median being
the robust for the situation. The places where values are less than
median can be used to identify observations which are having kind of
outlier behaviour within the dataset. Following equation gives us the
formula of getting segregated instances in \(M_v^k\).


\begin{figure}
\centering
\includegraphics{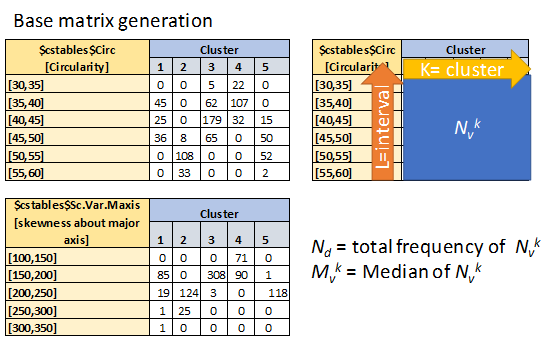}
\caption{Crosstab Base Matrix for 2 variable examples}
\includegraphics{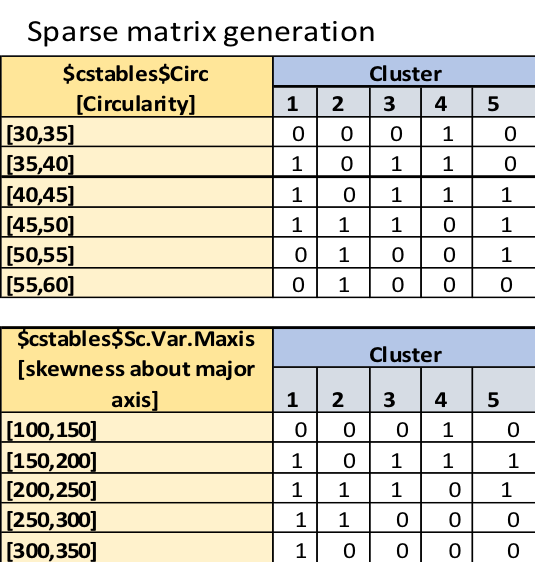}
\caption{Crosstab Sparse Matrix for 2 variable examples}
\end{figure}

\quad
\quad
\quad
\quad
\begin{equation}
N_v^k = \displaystyle \sum_{x= 1}^{l \ast k} \begin{cases}
1 & \text{ if } x > median(M_v^k) \\ 
0 & \text{ if } x <= median(M_v^k)
\end{cases}
\label{eq:weigthed1}
\end{equation}

As for \(explaination\ factor\), \(N_d\) observations has to be filled
in \(l \ast k\) places. The log ration of this defines how much of data
can be eplained by this combination of clusters and variables.

Since unique values of varible is being used, in our experiments taking
histogram of continuous varibles gave better results. Hence it is
adviced to be use histogram bucketing for calculating cluster score for
continuous variables.

Summation of \(S_v^k\) for all variables gives us the score of cluster
method on the data. As one can observe both parts of eqn. (1) are
monotonous in nature hence the multiplication wil also lead to
monotonous behaviour. This is can be well observed in experiments and
results section.

\hypertarget{procedure}{%
\section{4. Procedure}\label{procedure}}

The \(Algorithm 1\) describes the procedure for calculating scores for
the clusters

\textbf{\_\_\_\_\_\_\_\_\_\_\_\_\_\_\_\_\_\_\_\_\_\_\_\_\_\_\_\_\_\_\_\_\_\_\_\_\_\_\_\_\_\_}

\textbf{Algorithm 1:}: Cluster Score

\textbf{\_\_\_\_\_\_\_\_\_\_\_\_\_\_\_\_\_\_\_\_\_\_\_\_\_\_\_\_\_\_\_\_\_\_\_\_\_\_\_\_\_\_}

\textbf{Data\ }: \(T_{d}\) is training data with \(M\) variables and
\(N_d\) observtions,

\textbf{Inputs}: \(k\) number of clusters

\textbf{Output}: \(Cluster\ Score\ S^k\)

\textbf{Begin}

\qquad{\textbf{for} $v$ in $M$}

\qquad\quad{\textbf{if} $v$ is Numeric}

\qquad\quad\quad{ ${v}' = histogram(v)$ or $decile(v)$}

\qquad\quad\quad{$\mathbf{M_v^k}\ =$ Cross tab of $v'$ with $k$ clusters }

\qquad\quad\quad{$\mathbf{N_v^k} :=$ Compute segregated instances using $M_v^k$ from Eqn. (2) }

\qquad\quad{\textbf{else if} $v$ is Categorical / Ordinal}

\qquad\quad\quad{$\mathbf{M_v^k}\ =$ Cross tab of $v$ with $k$ clusters }

\qquad\quad\quad{$\mathbf{N_v^k} :=$ Compute segregated instances using $M_v^k$ from Eqn. (2)}

\qquad\quad{\textbf{end if;}}

\qquad\quad{$\mathbf{S_v^k} :=$ Compute cluster metric using $N_v^k,\ l,\ k,\ N_d$ from Eqn. (1) }

\qquad{\textbf{end;} }

\qquad{$\mathbf{S^k} = \sum_{v=1}^{M} S_v^k$ }

\textbf{End}

\textbf{\_\_\_\_\_\_\_\_\_\_\_\_\_\_\_\_\_\_\_\_\_\_\_\_\_\_\_\_\_\_\_\_\_\_\_\_\_\_\_\_\_\_}

\qquad

Since this method is non-parametric in nature, it is capable to
understand the internal boundaries which are drawn from different
cluster sizes and methods. Individual variable-cluster score Svk can be
used to assess the quality of segregation made by cluster on that
variable. Hence for drawing conclusion about clusters, one can almost
always look for variables with low scores. Owing to this property this
also helps in finding influential variables without using dependent /
target variable.

This formulation adheres to all the points mentioned to be a ``good''
\(metric\), as explained below:

\begin{enumerate}
\def\labelenumi{\arabic{enumi})}
\tightlist
\item
  The \(metric\) calculation is invariant of datatype.
\item
  Individual \(S_m^k\) can be used to see which variable is being
  explained more than others for finding similarity.
\item
  We can calculate same score for all variables that are not used for
  clustering and since this is not dependent on data, it is cross
  comparable.
\item
  For the \(metric\), order dose not matter. Hence makes it viable even
  if the ``name'' of cluster changes.
\item
  The \(metric\) makes no assumption about underlying measures and
  hence, is invariant for different clustering methods.
\item
  Owing to cross-tab behaviour the repeatability is assured, as the data
  is not going to change / or need to be re-calculated based on cluster
  method.
\end{enumerate}

Also pertaining to business uses, practitioner can artificially weigh
each variable to deduce the net score for clusters. These weights just
need to be multiplied to individual scores to get the weighted scores.
Number of bins / breaks in histogram for numeric variable can be kept
constant or coarse bucketing can be used to bin each variable. In our
experiment we observed, coarse buckets for individual variable yields
better results.

One by product of this method is when \(M_v^k\) calculated, clusters and
observations with outlier behaviour can be extracted and its
cross-feature influence can be estimated. This might be very helpful in
analysis of fraud detection, anomaly detections and other rare event
occurrence problems.

\hypertarget{experminetal-results}{%
\section{5. Experminetal Results}\label{experminetal-results}}

Showcased the capability of the metric with reference to understanding
optimum number of cluster based on Vehicle Silhouettes data. We have run
experiments with other datasets as well and the results are really
encouraging and deterministic. If we look at Figure-3, for K-Means at
close to 6 cluster our metric showcase highest seperation and high
sparcity of results. Where as for PAM it comes to 5 Clusters as seen in
Figure - 4.

\quad

\begin{figure}
\centering
\includegraphics{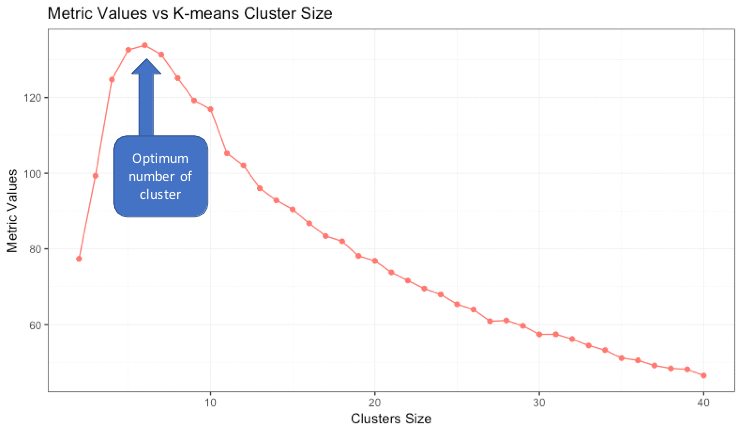}
\caption{Finding optimum Number of Cluster for K-mean Clustering on
Vehicle Data}
\end{figure}

Figuer-5 showcase the evaluation comparison of different clustering
techniques and how the metric value can help to compare them all with
common reference point.

\quad


\begin{figure}
\centering
\includegraphics{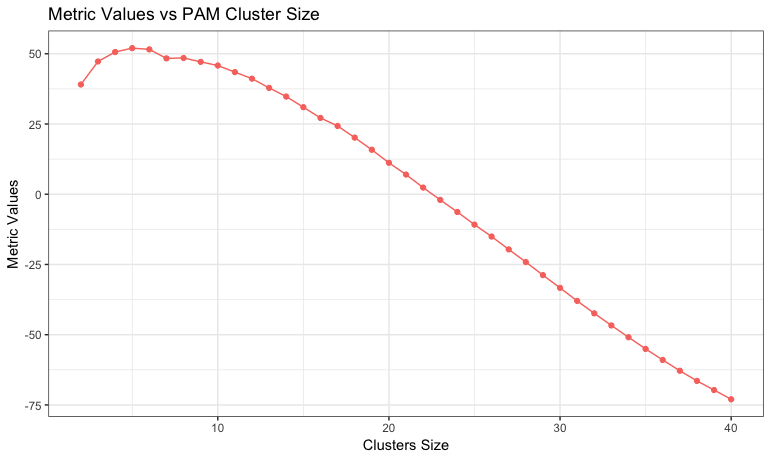}
\caption{Finding optimum Number of Cluster for K-mean Clustering on
Vehicle Data}
\includegraphics{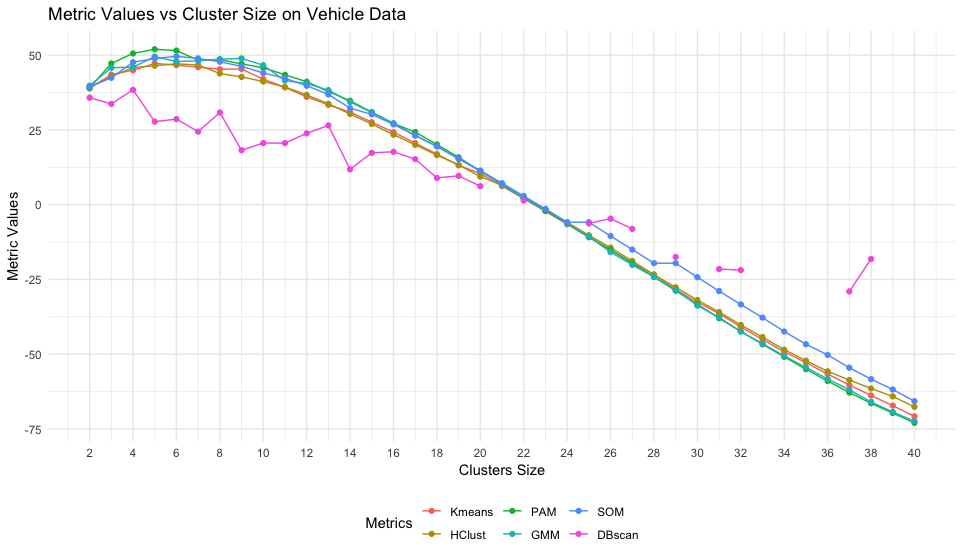}
\caption{Clustering Method comparison for Vehicle Data}
\end{figure}

As mentioned earlier our metric also helps to find th right features for
clustering. This is depicted on the Figure-6. Blue dots refers to the
metric value, Orange indicates the explanation factor and Red dots
represents the seggregation factors. This can help statistician to take
informend decision on the cluster health and what all variables can be
considered for clustering.

\quad

\begin{figure}
\centering
\includegraphics{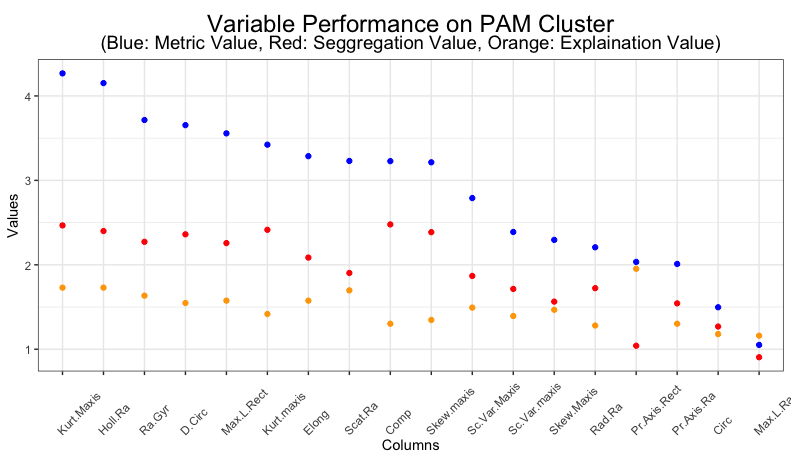}
\caption{Variable Impact of clusteris for Vehicle Data}
\end{figure}

Further analysis on other data set are depeicted on Appendix-1.

\hypertarget{conclusion-and-future-scope}{%
\section{6. Conclusion and Future
Scope}\label{conclusion-and-future-scope}}

In this paper we have proposed a new \(metric\) to estimate the cluster
behaviour from a non-parametric which is not based on any assumptions
about clustering method. We also showed how this \(metric\) can act as
tools for the statisticians for making more sense from the data. The
plots in our experiments not only helps in understand the cluster
behaviour but can also be very insightful in rare event modelling.

\quad

There is a possibility to explore and extend this \(metric\) in
classification and value estimation modelling. If applied in such
scenario, this can be implemented as loss function for linear and
non-linear methods so that proper segmentation of data can be achieved.

\hypertarget{references}{%
\section*{References}\label{references}}
\addcontentsline{toc}{section}{References}

\hypertarget{refs}{}
\leavevmode\hypertarget{ref-survey2}{}%
Berkhin, Pavel. 2002. ``Survey of Clustering Data Mining Techniques.''

\leavevmode\hypertarget{ref-cvi2}{}%
H. Chouikhi, N. Ghazzali, M. Charrad. 2015. ``A Comparison Study of
Clustering Validity Indices.'' \emph{Global Summit on Computer
Information Technology (GSCIT) and IEEE}, 1--4.
\url{https://doi.org/10.1109/GSCIT.2015.7353330}.

\leavevmode\hypertarget{ref-cvi3}{}%
H. Meroufel, N. Farhi, H. Mahi. 2017. ``Comparative Study Between
Validity Indices to Obtain the Optimal Cluster.'' \emph{International
Journal of Computer Electrical Engineering} 9 (1): 1--8.
\url{https://doi.org/10.17706/ijcee.2017.9.1.343-350}.

\leavevmode\hypertarget{ref-cvi4}{}%
J. Hamalainen, T. Karkkainen, S. Jauhiainen. 2017. ``Comparison of
Internal Clustering Validation Indices for Prototype-Based Clustering.''
\emph{MDPI Algorithms Articles}, 1--14.
\url{https://doi.org/10.3390/a10030105}.

\leavevmode\hypertarget{ref-pbm}{}%
M. K. Pakhira, U. Maulik, S. Bandyopadhyay. 2004. ``Validity Index for
Crisp and Fuzzy Clusters.'' \emph{Pattern Recognition} 37 (3): 487--501.
\url{https://doi.org/10.1016/j.patcog.2003.06.005}.

\leavevmode\hypertarget{ref-cvi1}{}%
O. Arbelaitz, J. Muguerza, I. Gurrutxaga. 2013. ``An Extensive
Comparative Study of Cluster Validity Indices.'' \emph{Pattern
Recognition} 46 (1): 243--56.
\url{https://doi.org/10.1016/j.patcog.2012.07.021}.

\leavevmode\hypertarget{ref-survey1}{}%
Xu, Rui, and D. Wunsch. 2005. ``Survey of Clustering Algorithms.''
\emph{IEEE Transactions on Neural Networks} 16 (3): 645--78.
\url{https://doi.org/10.1109/TNN.2005.845141}.

\quad

\quad

\quad

\quad

\quad

\quad

\appendix
\hypertarget{appendix-1-other-analysis}{%
\section{Appendix: Analysis on other Standard Data}\label{appendix-1-other-analysis}}

\begin{figure}
\centering
\includegraphics{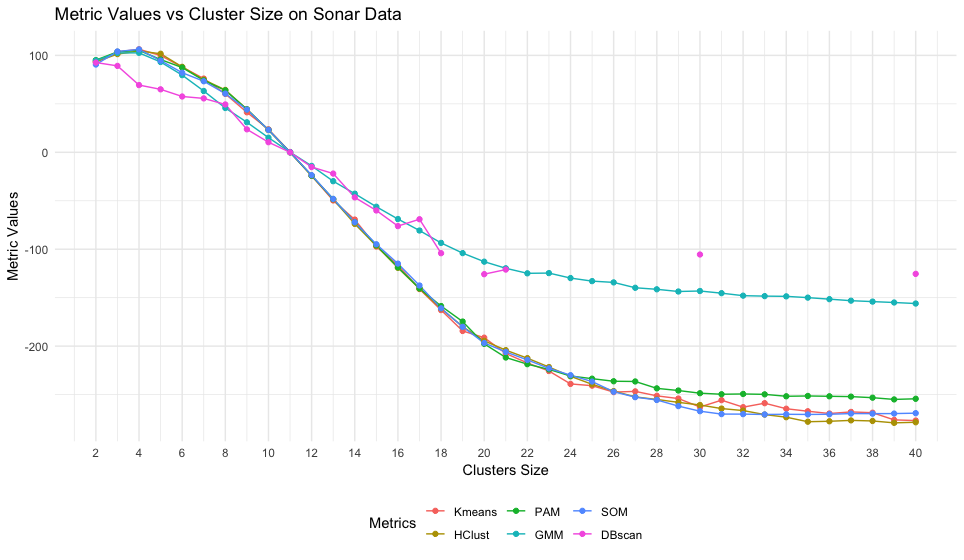}
\caption{Clustering Method comparison for Sonar Data}
\includegraphics{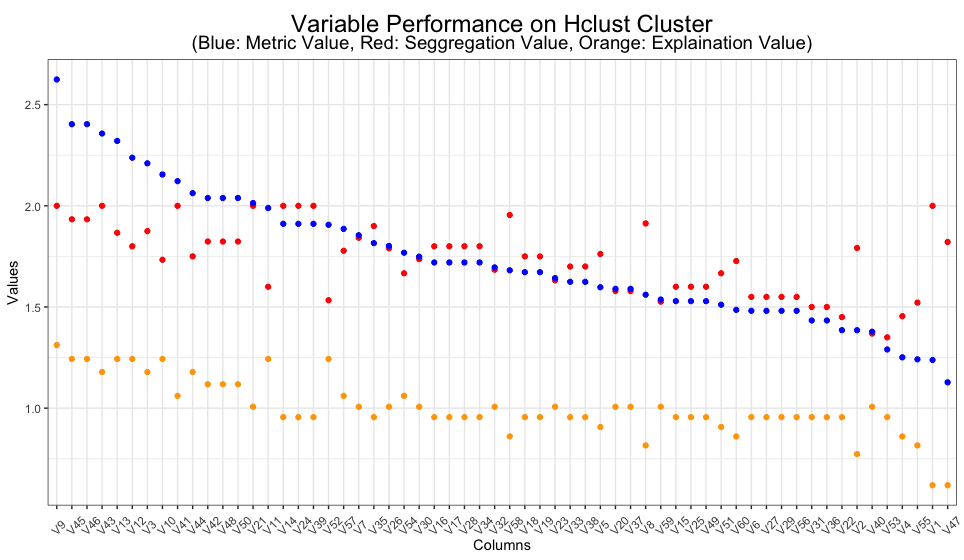}
\caption{Variable Impact of clusteris for Sonar Data}
\end{figure}

\begin{figure}
\centering
\includegraphics{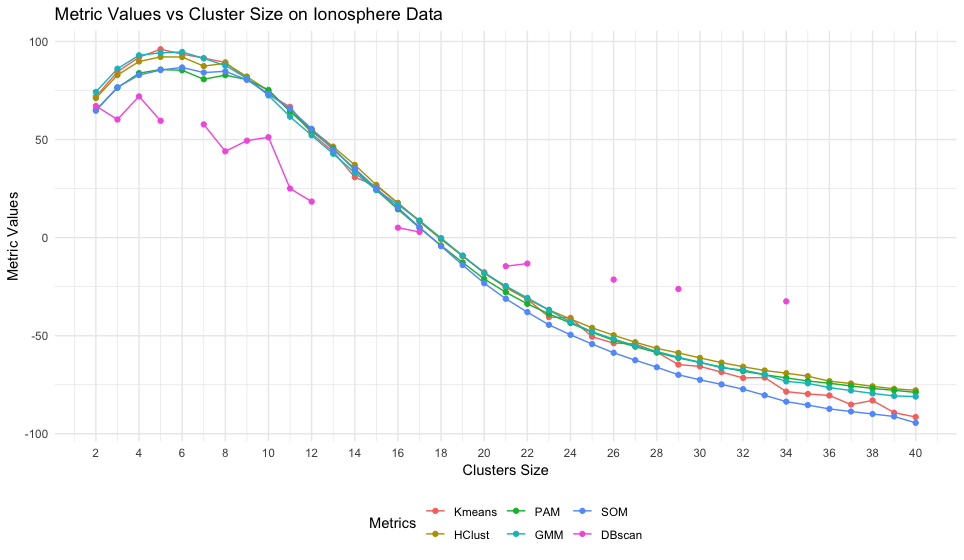}
\caption{Clustering Method comparison for Ionosphere Data}
\includegraphics{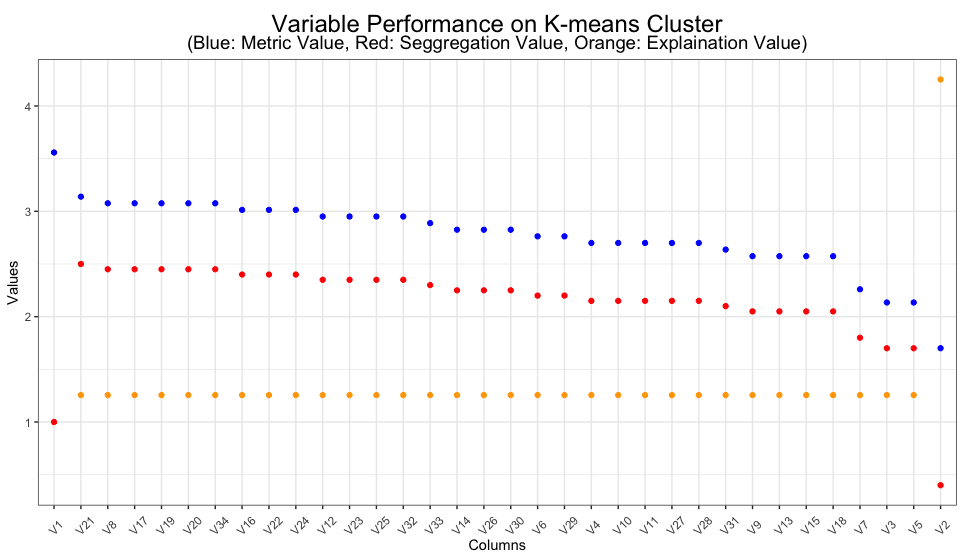}
\caption{Variable Impact of clusteris for ionosphere Data}
\end{figure}

\begin{figure}
\centering
\includegraphics{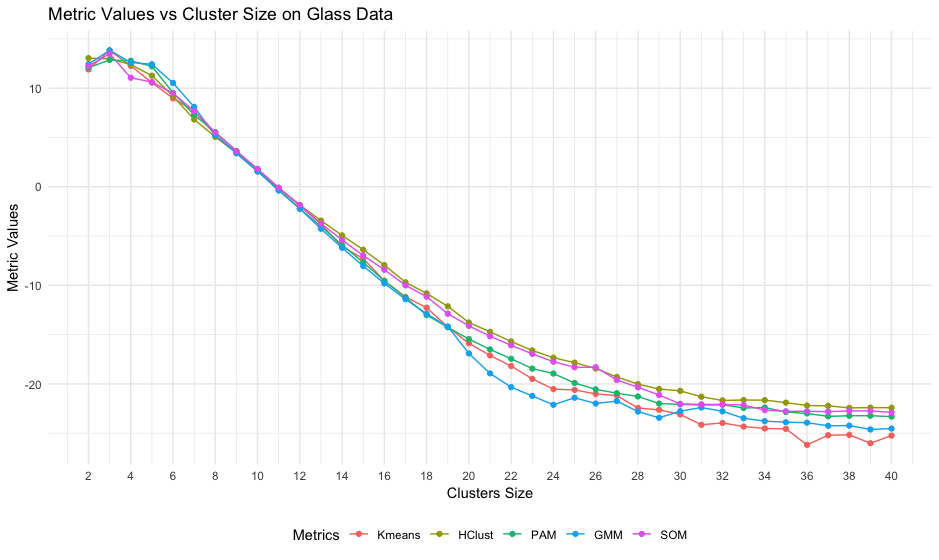}
\caption{Clustering Method comparison for Glass Data}
\includegraphics{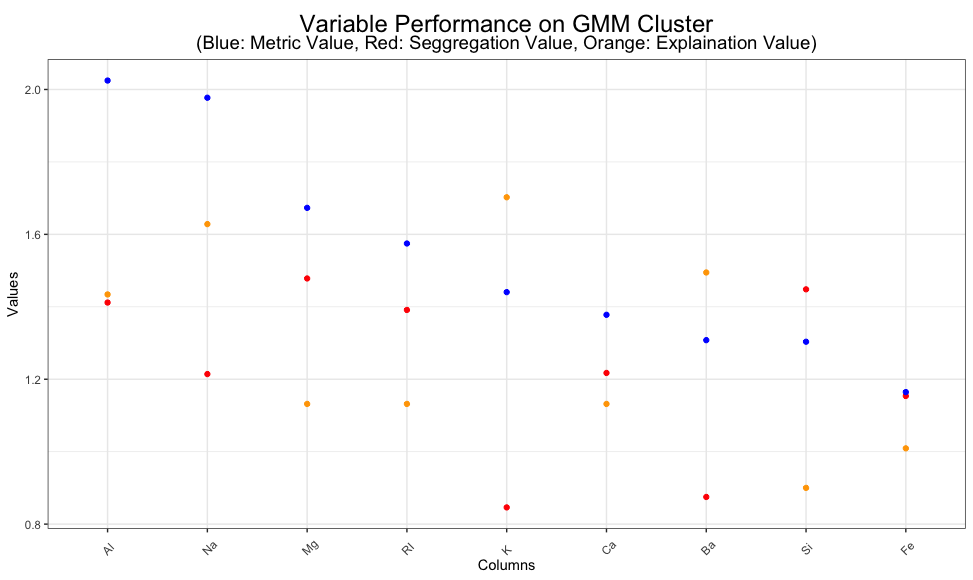}
\caption{Variable Impact of clusteris for ionosphere Data}
\end{figure}

\end{document}